# Automating Sound Change Prediction for Phylogenetic Inference: A Tukanoan Case Study

**Kalvin Chang**[1*]  **Nathaniel R. Robinson**[1,2*]  **Anna Cai**[1*]
**Ting Chen**[1]  **Annie Zhang**[1]  **David R. Mortensen**[1]

[1]School of Computer Science, Carnegie Mellon University
[2]Center for Language and Speech Processing, Johns Hopkins University
kalvinc@alumni.cmu.edu, nrobin38@jhu.edu,
{annacai, tingc2, ruoxinz, dmortens}@cs.cmu.edu

## Abstract

We describe a set of new methods to partially automate linguistic phylogenetic inference given (1) cognate sets with their respective protoforms and sound laws, (2) a mapping from phones to their articulatory features and (3) a typological database of sound changes. We train a neural network on these sound change data to weight articulatory distances between phones and predict intermediate sound change steps between historical protoforms and their modern descendants, replacing a linguistic expert in part of a parsimony-based phylogenetic inference algorithm. In our best experiments on Tukanoan languages, this method produces trees with a Generalized Quartet Distance of 0.12 from a tree that used expert annotations, a significant improvement over other semi-automated baselines. We discuss potential benefits and drawbacks to our neural approach and parsimony-based tree prediction. We also experiment with a minimal generalization learner for automatic sound law induction, finding it comparably effective to sound laws from expert annotation. Our code is publicly available.[1]

## 1 Introduction

Languages and biological species evolve in interestingly analogous ways. Both display variation in space and time that may be inherited or innovated. As in biology, each node in a linguistic phylogenetic (family) tree corresponds to one or more innovations ("mutations"). Typically, linguists infer these phylogenies by finding patterns of innovations in pronunciation, or sound changes.

SOUND LAWS, or rules that define sound changes and the contexts in which they occur, apply to all instances of a sound in a given context. (For example, all instances of Proto-West Germanic [*t] became [t͡s] (written as z) at the beginning of words in High German, while English was unaffected. This resulted in the English-German cognate pairs *zehn* : ten, *Zoll* : toll, and *Zahn* : tooth.) Because these laws have few exceptions, they can work as a basis for modeling historical language change. Linguists typically infer phylogenies by constructing the tree that maps from the ancestor language at the root to the daughter languages via the most probable system of these sound laws (Hoenigswald, 1960). Existing partially automated approaches to this method require multiple sets of expert annotations. We attempt to alleviate this via proposed methods that incorporate even more automation. Below we discuss the necessity of both sound laws and sound changes in predicting phylogenies, which we automatically infer in our proposed methods.

Linguists induce sound laws by aligning COGNATES (words with a common ancestor) by phoneme. From this alignment they extract SOUND CORRESPONDENCES, or sets of sounds in the same context that likely evolved from the same sound in the proto-language. (For example, at the beginning of words, there is a correspondence between High German [t͡s], Dutch [t], English [t], Swedish [t], and Icelandic [t]). They then reconstruct protophonemes for each set of aligned cognates (in our Germanic example, this happens to be [*t]). The posited sound laws from this process enable deterministic derivation of the daughter forms from the reconstructed PROTOFORMS, or words in the proto-language. Inducing sound laws is central to sound change-based phylogenetics. We experiment with both algorithms that predict these sound laws automatically and those that need them to be provided by a linguist. Beyond sound laws alone, however, phylogenetic inference algorithms must consider how sounds evolve and branch off through INTERMEDIATE SOUND CHANGES over time.

---
[*]Authors contributed equally.
[1]https://github.com/cmu-llab/aiscp

Sound change emerges from phonetic variation as speakers modify their pronunciation along acoustic or articulatory dimensions (Garrett et al., 2015; Garrett and Johnson, 2013; Lindblom et al., 1995). Because some variations in pronunciation occur more frequently than others, not all sound changes are equally probable. In particular, the probability of a sound change (e.g. [p] becoming [f]) is often different from that of its reverse (e.g. [f] becoming [p]), a property known as the DIRECTIONALITY of sound change (Campbell, 2013; Chacon and List, 2016). Because phonetic variation is gradual (with few ARTICULATORY FEATURES—or fundamental characteristics of pronunciation—changing at a time) (Sievers, 1901; Brugmann and Osthoff, 1878; Paul, 2010), sound change often results in phonetically similar sounds across cognates and their corresponding protoform. Larger apparent jumps in pronunciation from PROTO-PHONEME (ancestral sound) to REFLEX (descendant sound) are often the result of smaller changes over time, or intermediate sound changes (Garrett et al., 2015; Beguš, 2016). For example, k > t͡ʃ ([k] becomes [t͡ʃ]) may encompass the chain of sound changes k > kʲ > c > t͡ʃ. Intermediate paths from a proto-sound to different daughter reflexes can overlap, which enables identifying innovations that are shared among daughters (SHARED INNOVATIONS).

## 1.1 Contribution

We automate portions of Chacon and List (2016)'s phylogenetic inference method via our novel Automatic Intermediate Sound Change Prediction (AISCP) method and experiment with further automation via Automatic Sound Law Induction (ASLI).

Chacon and List (2016) rely on expert judgements for Tukanoan sound changes, which we replace at different stages of their algorithm. Our main contribution is replacing expert-provided intermediate sound changes with AISCP— essentially "invent[ing]" proto-sounds not seen in reflexes, which many unsupervised protoform reconstruction models cannot do (List, 2022). These AISCP predictions rely on (1) a PHONOLOGICAL PRIOR based on articulatory distances (Mortensen et al., 2016) and (2) TYPOLOGICAL GROUNDING learned by a neural network from a database of multilingual sound changes. The phonological prior captures the tendency for sounds to change into sounds that are pronounced similarly, while the typological grounding encodes the direction and frequency of sound changes. Our results show that phylogenetic inference with AISCP approaches expert performance in a computational paradigm requiring expert knowledge only for cognate sets, sound laws, and protoforms.

In additional experiments, we further automate the process via ASLI: predicting not just intermediate sound changes, but sound laws from protoforms and reflexes. We induce these laws via methods from Albright and Hayes (2003) and Wilson and Li (2021), newly applied for ASLI.

We conduct experiments on data from Tukanoan languages, spoken in Columbia, Brazil, Peru, and Ecuador. The data contain Proto-Tukanoan reconstructions from a leading Tukanoan linguist, Chacon (2013, 2014). We take their reconstruction and sound changes as our gold standard, as did Chacon and List (2016). In summary, we contribute:

1. A training paradigm by which a neural network can produce phonetically natural intermediate sound changes as a typological grounding for AISCP
2. Experimental evidence that AISCP can approach expert phylogenetic inference, with automatic correct groupings of West Tukano and East-Eastern Tukano
3. Ablations indicating that intermediate sound changes and directional weighted sound transition costs are useful to predict phylogenies
4. An ASLI method for phylogenetic inference that achieves comparable performance to expert-provided sound laws when combined with AISCP for maximal automation
5. Analysis suggesting parsimony-based phylogenetic inference may be unreliable

## 2 Related work

Unlike our work, prior phylolinguistic work mostly inferred a tree from a boolean cognacy matrix that shows which synonymous words come from the same ancestral word (Greenhill et al., 2020). However, cognacy is complicated by language contact that leads to the borrowing of words, as opposed to their inheritance (Ryskina et al., 2020; Francis et al., 2021). Campbell (2013) criticized such use of cognacy information in phylogenetic inference, and called on computational methods to use shared innovations as linguists do. Zheng (2018) heeded this call and manually derived shared innovations for Proto-Min and its modern daughters, finally running a maximum parsimony algorithm from Felsenstein (2013) on these shared innovations. However, their shared innovation matrix is binary and does not encode the direction and frequency of sound changes as our methods do.

Hruschka et al. (2015) jointly inferred phylogeny and reconstructed protoforms with Markov Chain Monte Carlo (MCMC) using phonological data from Turkic languages, where the tree likelihood was conditioned on reconstructed protoforms. However, their sound laws are all context-free. Clarté and Ryder (2022) perform joint phylogenetic, protoform reconstruction, and cognate inference for 14 Polynesian languages using MCMC, but the expressive power of their model is limited to only CVCV sequences for alignment without insertions or deletions or sound law contexts. Unlike these approaches, our methods (both with and without ASLI) include in-context sound laws and can process all sound sequences.

We are also not the first researchers to explore neural modeling of phonetic features. Hartmann (2019, 2021) showed that neural networks can predict features of Proto-Indo-European phones given the features of a trigram context, which reflect SYNCHRONIC (applying at a particular stage in a language's history) phonetic phenomena. Our neural network, on the other hand, predicts the probability of feature changes in a sound change, given DIACHRONIC data (data that represents change over time).

In recent years, there has also been existing work on ASLI. Luo (2021) used reinforcement learning with hierarchical Monte Carlo tree search to induce sound laws for Germanic, Romance, and Slavic. To our knowledge, they are the first to propose an ASLI method. List (2019) also attempted automatic induction of sound correspondences, though these correspondences lacked the contexts associated with actual sound laws. The minimum generalization learner we employ for ASLI (Albright and Hayes, 2003), in contrast, was originally designed to induce synchronic morphological rules and is deterministic.

## 3 Methodology

We incorporate AISCP and ASLI into Chacon and List (2016)'s DiWeST method for phylogenetic inference, which involves directed, weighted phone transitions. The authors describe it as a directed version of Sankoff parsimony (Sankoff, 1975). Their maximum parsimony algorithm searches for trees by trying different sound transitions (intermediate sound changes) along the branches of trees that minimize the total transition cost. By making the cost of sound changes asymmetrical in a sound change transition matrix, their method captures the directionality of sound change, yielding a rooted tree. To search the large space of possible trees, they used a genetic search algorithm that balances exploration (iterating through a subset of possible trees) and exploitation (incrementally mutating the current best trees). Chacon and List (2016)'s algorithm follows this framework:

1. Align protoforms with reflexes (performed by an expert)
2. Learn sound laws between protoforms and reflexes (performed by an expert)
3. Create a sound change transition matrix (largely performed by an expert)
    (a) Identify intermediate sound changes (Section 3.2)
    (b) Assign a weight to intermediate sound change transitions (Section 3.2.1)
4. Perform maximum parsimony-based phylogenetic inference using the transition matrix from the step above
5. Obtain a consensus tree

Our algorithm follows this same framework. Our contribution is to automate the expert's annotations in step 3, and in steps 1-2 in some experiments. Section 3.1 outlines the way both Chacon and List (2016)'s algorithm and our modifications of it incorporate intermediate sound changes via a sound transition matrix. Section 3.2 elaborates on replacing the expert in step 3 with AISCP. Section 3.3 outlines replacing the expert in steps 1-2 with ASLI.

### 3.1 Creating the sound change transition matrix

Chacon and List (2016)'s transition matrix specifies the cost of intermediate sound changes that the parsimony algorithm tries. It is constructed by creating a directed graph (with phones as nodes) of the possible intermediate sound changes given by a linguist for each sound correspondence.[2] (In this context, we use the term CORRESPONDENCE to mean a proto-phoneme and its reflexes.) The ex-pert may identify more than one possible path of intermediate phones between the proto-phoneme and reflex (e.g. k > kʲ > c > t͡ʃ and k > x > h > ʃ > t͡ʃ). The transition cost from one phone (source) to another (target) on an intermediate path is simply the length (in edges) of the shortest path from source to target in the directed graph for the correspondence. This value is encoded in the transition matrix at the row corresponding to the source index and the column corresponding to the target index. Pairs of source and target phones with no connecting path in the graph are penalized with a high transition cost. Paths are all directed from the proto-phoneme towards the reflex, encoding sound change directionality. Our AISCP algorithm's transition matrix is also directional, but the intermediate sound changes and edge weights are derived from the probability of articulatory features changing, as predicted by a neural network.

---

[2]See Figure 5 in Chacon and List (2016) for a diagram of the process.

## 3.2 Automatic intermediate sound change prediction (AISCP)

To automate intermediate sound change prediction, we create a fully connected graph using a mapping $x$ that encodes each phone $p$ as a ternary vector of $N$ articulatory features (such as [voice] or [syllabic]), where each feature in position $f$ is encoded as $-1$ (not present), 0 (not applicable), or 1 (present); $x(s)_f \in \{-1, 0, 1\}$. This encoding lets us consider information shared by phones. Encodings for [d] and [t] differ in only one articulatory feature: $x([d])_{[voice]} = 1$, while $x([t])_{[voice]} = -1$. We can quantify such phonetic similarity between sounds using Mortensen et al. (2016)'s feature edit distance (FED). FED is Levenshtein edit distance, where the cost of an edit is the proportion of articulatory features changed. It reflects phonetic similarity between sounds: FED([t], [k]) has four times the value of FED([t], [d]), since the former pair requires four feature edits. For our phonological prior, we create a graph where the nodes are IPA phones, and all node pairs are joined by an undirected edge with weight equal to the FED. In this graph, intermediate phones are interpreted as the phones on the least-weighted paths between the proto-sound and the reflex. There can be multiple least-weighted paths between nodes, just as there can be multiple transition paths in a correspondence.

### 3.2.1 Neurally weighted FED for AISCP

The way we modify FED is central to our approach. FED has undesirable traits for modeling sound change — it is not directional (e.g. there is no way to encode whether p > f or f > p is more likely), and it gives an equal cost to every feature change, regardless of the source phone. This neglects information about sound change tendencies: for example, [d] is more likely to change its [voice] feature and become [t] than to change its [sonorant] feature and become a sonorant (like [l] or [r]).

We propose directional weighted feature edit distance (DWFED) to model these realities, by training a neural network to predict the cost of each feature change, given the source phone. The network learns each feature's *directional* change costs: i.e. the cost of the [voice] feature increasing (*voicing*) may differ from the cost of [voice] decreasing (*devoicing*). We interpret a feature edit's cost as one minus the probability of its occurring. Thus we train the neural network to model the *probability* of each directional feature edit, conditioned on the source phone, e.g. $P(voicing \mid source = [p])$.

The network predicts this for all articulatory features. It uses the encoding function $x$ described in Section 3.2 to encode each source phone $s$ as a vector of feature values in positions $f$. For an arbitrary target phone $t$, the network predicts both probabilities $P(x(t)_f > x(s)_f \mid s)$ and $P(x(t)_f < x(s)_f \mid s)$, for each $s$ and $f$. (We write these in shorthand as $P(f \uparrow)$ and $P(f \downarrow)$, respectively.) It does this by learning the mapping $M : \{0, 1\}^{3N} \rightarrow \{0, 1\}^{2N}$, where $M(v) = \sigma(\text{NN}(v))$, a series of linear layers with ReLU activations followed by a sigmoid activation at the end. The input is a one-hot encoding of the source phone's $N$ articulatory features (each having value $-1$, 0, or 1), resulting in a binary vector of length $3N$. The output contains the two directional sound change probabilities mentioned, for each of the $N$ features $f$, resulting in a length-$2N$ vector with values between 0 and 1.

Hence for a single-layer neural network, the model weights are a single $2N \times 3N$ matrix, where each entry encodes a feature's importance in determining another feature's probability of increasing or decreasing. For example, if $N = 24$, the weight matrix's [0,47] entry encodes the importance of the source phone's first feature equaling $-1$ in predicting whether its 24th feature will decrease. This also ensures some continuity: source phones with many common features will elicit similar output probabilities. However, because a feature value's probability of increasing or decreasing may depend on a combination of the source phone's features, we experiment with deeper neural networks to capture more intricate relationships (Section 4.2).

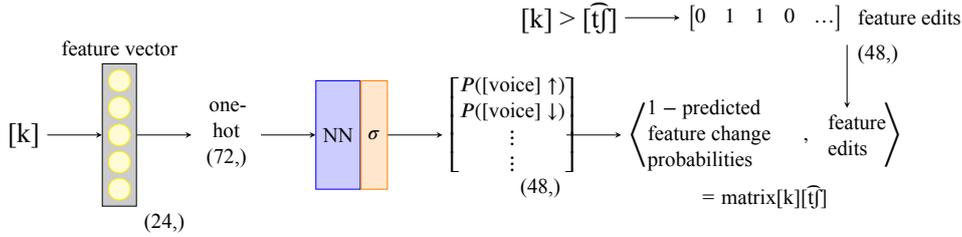

Figure 1: Calculation of the edge weights for the phone transition graph via DWFED

These networks can be trained on a database of real sound changes (the typological grounding mentioned in §1.1), by converting each sound change with source phone $s$ and target $t$ into a length-$3N$ binary vector encoded from $s$ and one length-$2N$ binary vector representing the ground-truth sound change direction (i.e. whether each feature increased and whether it decreased from $s \rightarrow t$). We use the length-$3N$ vector as the network's input and the length-$2N$ vector to compute loss with its output. In this way, the neural network learns which features tend to change in which directions for each source phone in natural languages. Refer to Figure 1 for a diagram of our method in the case $N = 24$. Since the length-$2N$ vector representing a direction of sound change is binary, using the dot product to multiply it with the output of the neural network at inference time extracts the relevant probabilities needed to calculate the DWFED of transitioning from a source phone to a target. (The "matrix" mentioned in the figure refers to the connected phone graph whose edge weights are determined by DWFED.)

### 3.3 Automatic sound law induction (ASLI)

In addition to automating step 3 of the algorithm in Section 3 via AISCP, we experiment with ASLI via a minimal generalization learner from Wilson and Li (2021), instead of using sound laws from an expert in steps 1 and 2. This can be done by aligning the phones in protoforms and daughters (still provided by an expert) via Needleman-Wunsch alignment (Needleman and Wunsch, 1970), a Levenshtein edit distance alignment algorithm used in computational biology. We adapt the algorithm so that the substitution cost between two phones is the FED rather than a constant. This ensures that similar phones like [t] and [d] will align, rather than more distant phones like [t] and [k]. See Figure 2 for an example of our *alignment* process.

Our ASLI method uses Albright and Hayes (2002, 2003)'s *minimal generalization* algorithm, as adapted by Wilson and Li (2021). These methods were developed for synchronic sound rules. However, since such rules reflect sound changes (Ohala, 2003), we repurpose the method for diachronic sound laws. Albright and Hayes generate the base rules by taking the longest common prefix and longest common suffix from each word pair as the context and treating the remaining strings as a rule, then iteratively generalizing the set of rules based on shared contexts. Because sound changes usually involve individual phones, we generate a base rule for every phone-level change in the aligned protoform and daughter instead. The *rule induction* process in Figure 2 shows sound law extraction, prior to iterative generalization.

## 4 Experiments

### 4.1 Dataset

In all our experiments we used Tukanoan expert annotations from two sources. For AISCP (without ASLI), we use expert-provided sound laws from 33 sound correspondences for 21 Tukanoan languages from Chacon and List (2016). Unfortunately, the phonological and lexical data needed for alignment and ASLI is not available for all 21 varieties. Thus for our ASLI experiment (Section 3.3) we used Chacon (2014)'s dataset of 15 Tukanoan languages.[3] This version contains phonetic transcriptions of daughters, cognacy, expert alignment (for manual evaluation and debugging), and reconstructed protoforms for 149 cognate sets, totalling 1,542 entries.

---

[3]https://github.com/lexibank/chacontukanoan/

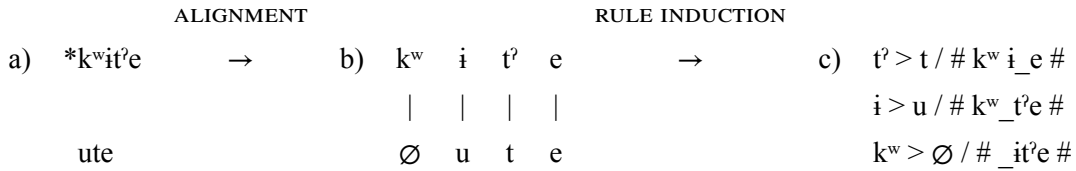

Figure 2: Alignment and induction of sound changes. A protoform and a daughter form are aligned, allowing the induction of base sound laws that can then be iteratively generalized using the minimal generalization learner.

### 4.2 Implementation

As outlined in Section 3.2, our AISCP method consists of (1) encoding phones as vectors of features and (2) training a neural network to calculate DWFED between the feature vectors, to weight the transition edges in a phone graph. We used PanPhon (Mortensen et al., 2016) as our mapping $x$ from phones to $N = 24$ articulatory features. To compute DWFED, we trained neural networks on 7,042 sound changes in multiple language families[4] from *Index Diachronica*[5] (Anonymous, 2016). Training on actual sound correspondences collected by linguists ensures that the model learns the directionality of sound change. Because PanPhon has $N = 24$ articulatory features, all our neural networks accept length-72 vectors as input and output length-48 sound change probability vectors. We used Binary Cross Entropy Loss since the reference length-48 sound change vectors are binary. We trained multilayer perceptrons of differing depths: 1 layer, 4 layers, 8 layers, and 16 layers (the latter two with skip connections).

Using neural DWFED we produce a phone graph containing a subset of the phones supported by PanPhon.[6] We include the null phone ∅ to model insertions and deletions in sound changes. These we penalize with a cost multiplier (15 for insertions and 10 for deletions), since substitutions are more common along intermediate paths. We find the graph's shortest paths using NetworkX (Hagberg et al., 2023) to produce intermediate paths for the sound transition matrix of each correspondence in the data.

For our experiments with ASLI (as outlined in Section 3.3), we (1) align protoforms with daughters via Needleman-Wunsch alignment, (2) extract sound laws as in Figure 2, and (3) perform iterative generalization. We modified FED slightly for our alignment: we penalized substitutions between vowels ([+syl, -cons]) and non-syllabic consonants ([-syl, +cons]) to prevent unnatural substitutions. We also filtered out sound laws with raw accuracy ≤ 0.6 before iterative generalization, since our approach to single-phone law extraction generated superfluous laws otherwise.

When running the parsimony-based algorithm, we searched through 10,000 trees, though Chacon and List (2016) found that most best trees can be found within the first 5,000. When the algorithm resulted in multiple trees with the same score, we obtained a consensus tree with the `consense` program from PHYLIP (Felsenstein, 2013).

We also included two ablation experiments: *standard FED* and *direct paths*. Our *standard FED* ablation consists of applying AISCP with standard FED, rather than neural DWFED. Although FED is not directional and treats all articulatory features as the same, it does not preclude the directionality of sound change in our algorithm, since the sound change matrix encodes direction by only considering paths that lead from the proto-phoneme to the reflex (as mentioned in Section 3.1). Our *direct paths* ablation does not use AISCP or expert-provided intermediate sound changes. It uses sound laws from the Tukanoan expert directly without any intermediate sound changes. We performed this experiment to probe whether intermediate paths are necessary for the inference algorithm. In this ablation, we also used standard FED as the weight of transition directly from proto-phonemes to reflexes.

### 4.3 Baselines

We compare our adaptations of Chacon and List (2016)'s algorithm (Section 3) using AISCP and ASLI to two baseline inference methods: *cog-*

---

[4]We manually removed Altaic sound correspondences from the database, since the proposed family is controversial.

[5]We chose *Index Diachronica* instead of the more authoritative KonsonantalWandel (Kümmel, 2007) because the latter lacks a public LaTeX source and only includes consonants.

[6]To ensure matrix curation was computationally tractable, we excluded all phones with diacritics other than those marking length, aspiration, and glottalization.

*nacy* and *shared innovations*. Both of these baselines use undirected binary matrices for parsimony-based phylogenetic inference. The first uses a cognacy matrix, indicating simply which daughter languages have entries for which cognate sets Chacon (2014). This is commonly used in phylogenetic inference (Greenhill et al., 2020), but it only considers lexical innovations and not phonological innovations. The second baseline uses a shared innovation matrix that indicates which languages participate in each sound law, i.e. innovate on the proto-phoneme in the conditioning environment of the law For both baselines we used PHYLIP's `Penny` program (Felsenstein, 2013) for parsimony-based phylogenetic inference, as it accepts binary matrices more readily than Chacon and List (2016)'s method.

### 4.4 Evaluation

We take the consensus tree from Chacon and List (2016) as the gold tree, which is a consensus between their DiWeST phylogenetic inference and the tree from (Chacon, 2014). To measure the distance between the gold and predicted trees, we use Generalized Quartet Distance (GQD), which groups leaf nodes (daughter languages) into *stars* and *butterflies* (Pompei et al., 2011). A *star* is a group of four leaves such that the most recent common ancestor of any pair among them is also the most recent common ancestor of all four. A *butterfly* is any quartet of leaves that is not a *star*.[7] GQD is the difference between the number of butterflies in the gold tree and the number of shared butterflies in both hypothesis and gold, normalized by the number in the gold. Because it does not penalize *stars*, GQD is well-suited to non-binary trees such as phylogenies (where *stars* persist, barring enough evidence to binarize them) (Sand et al., 2013; Pompei et al., 2011; Rama et al., 2018). For each experiment we report the minimum and mean GQD across ten runs of 10,000 trees each (which we found comparable to searching 100,000 trees).

## 5 Results and Discussion

Table 1 shows all experimental results. The *cognacy* baseline diverged greatly from the gold tree, while the *shared innovations* baseline captured about two-thirds of the gold tree butterflies, affirming the usefulness of phonological information. Our best tree on all 21 languages reproduced 88.0% of gold butterflies, using AISCP with DWFED instead of an expert. (We discuss this tree in Section 5.1; see Figure 6.) Our best tree on a subset of 15 languages, reproducing 87.6% of gold butterflies, was inferred via the maximum automation possible: using both AISCP and ASLI.[8] (See Section 5.3.) Our *standard FED* ablation achieved worse mean GQD than any experiments using expert sound laws, suggesting DWFED is more effective in creating the sound transition matrix. The *direct paths* ablation performed worse than shallow networks but outperformed some deeper networks, indicating that intermediate sounds are generally useful, but the quality of the sounds matters. (See Section 5.2.)

Our findings suggest that parsimony does not correlate with GQD, with Spearman's $\rho = -0.04$ across AISCP-only experiments (Figure 7). The parsimony of our best tree overall was not even better than the median across the 10 runs of its experiment. It seems relying only on parsimony to predict phylogenies is not guaranteed—or perhaps even likely—to produce optimal trees.

The high variance across experiments is likely due to Chacon and List (2016)'s genetic search algorithm starting with random trees and at times getting stuck in sub-optimal areas of the search space.

### 5.1 Recovering major Tukanoan groupings

We analyze the best tree's recovery of subgroups proposed by Chacon (2014), which largely recur in the consensus tree from Chacon and List (2016). Our algorithm correctly groups "Western Tukano" and "East-Eastern Tukano" varieties into their respective subgroups but not "West-Eastern" Tukano. Within the correctly grouped subgroups, the relative chronology of the branch is incorrect. (See Appendix B for details.) Overall, the larger subgroups within Tukanoan are correctly displayed, showing that our method can capture broad phylogenetic relationships as a linguist would. Additionally, our parsimony method from Chacon and List (2016) produces binary trees with all language pairs split in an overly specific

---

[7]See https://cran.r-project.org/web/packages/Quartet/vignettes/Quartet-Distance.pdf for a visualization of the 3 possible arrangements of butterflies.

[8]Note that we cannot compare the GQD scores in rows 9-12 of Table 1 directly with the rest, since they were calculated for a subset of 15 Tukanoan languages, while rows 1-8 display results on all 21 Tukanoan languages in our set. See Section 4.1.

|    | Section | Experiment | GQD (Min) ↓ | GQD (Mean ±σ) ↓ |
|----|---------|------------|-------------|-----------------|
| 1  | §4.3    | Baseline: cognacy | 0.533 | 0.533 |
| 2  |         | Baseline: shared innovations | 0.355 | 0.355 |
| 3  | §4.2    | C+L, w/ AISCP (*standard FED* ablation) | 0.325 | 0.440 ±0.062 |
| 4  |         | C+L, w/ AISCP (*direct paths* ablation) | 0.281 | 0.397 ±0.072 |
| 5  | §3.2    | C+L w/ AISCP, 1 layer NN | **0.120** | **0.295** ±0.118 |
| 6  |         | C+L w/ AISCP, 4 layer NN | 0.191 | 0.309 ±0.096 |
| 7  |         | C+L w/ AISCP, 8 layer NN | 0.402 | 0.439 ±0.021 |
| 8  |         | C+L w/ AISCP, 16 layer NN | 0.248 | 0.435 ±0.080 |
| 9  | §3.3    | C+L w/ AISCP + ASLI 1 layer NN | **0.124** | **0.224** ±0.076 |
| 10 |         | C+L w/ AISCP + ASLI, 4 layer NN | 0.354 | 0.461 ±0.070 |
| 11 |         | C+L w/ AISCP + ASLI, 8 layer NN | 0.237 | 0.433 ±0.092 |
| 12 |         | C+L w/ AISCP + ASLI, 16 layer NN | 0.396 | 0.483 ±0.084 |

Table 1: Result of experiments across 10 runs. C+L refers to Chacon and List (2016)'s parsimony method outlined in (Section 3).

### 5.2 Analyzing AISCP's intermediate paths

Intermediate paths from the phone graph using our best performing, 1-layer network are phonetically and typologically natural. We predict *k > *c > *tɕ > t͡ʃ for proto-sound *k and reflex t͡ʃ, with [c] and [tɕ] not observed in the daughters but plausible as intermediate phones. Another predicted path is *p > *f > h, where [f] is unobserved; p > f appears 16 times in our subset of *Index Diachronica*, and f > h is acoustically motivated since [f] and [h] are both characterized by low-amplitude aperiodic nose. This shows the viability of using articulatory features to model phonetically motivated intermediate sound changes in future research. As a comparison, Chacon and List predicted *k > *kʲ > t͡ʃ (or *k > kʰ > t͡ʃ) and *p > *pʰ > *ɸ > h. (Note that they skipped a palatalization step or two in the former.) DWFED does not reproduce these expert paths perfectly, since it prefers paths matching feature change tendencies learned from *Index Diachronica*.

We find that many intermediate paths predicted in our *standard FED* ablation are also phonetically plausible, e.g. k > k͡x > t͡ɕ > t͡ʃ and j > ʒ > d͡ʒ > t͡ʃ. However, unweighted FED produces unreasonably many intermediate paths and many sound changes per path, resulting in phonetically unnatural paths, such as tʔ > dʔ > zʔ > ʔɾ > rʔ > r. For this same sound correspondence, our ablation includes all phonetic variants with the same FED, with no typological intuition. DWFED instead restricts the number of intermediate paths by favoring more typologically usual ones. While this results in plausible paths, DWFED yields only one unique intermediate path (Table 2) for each proto-phoneme and reflex pair. This is not entirely desirable, as proto-phonemes and reflexes may have multiple plausible paths. (The average number of paths in expert transition matrices is > 1; see Table 2.) The ideal setting is to include some of the most plausible paths, since this allows paths with higher DWFED that are in fact attested to be considered.

All neural approaches and the expert produce intermediate paths with an average of ∼ 2 edges (compared to 3.47 for our *standard FED* ablation.) The different networks also have similar expert sound change recall to each other. Thus their ability to replicate the length or phones of the expert intermediate paths cannot explain the ∼ 0.1 difference in GQD between the shallower and the deeper networks. (Indeed, the *standard FED* ablation has higher recall but performs worse.) This suggests that simply replicating the expert's intermediate paths is not sufficient without correctly reflecting the relative weights of the sound changes.

A naive alternative to weighting edit distances neurally is down-weighting the absolute FED between phones for attested sound changes. Our neural approach, however, is preferable. The naive approach is analogous to connecting cities (phones) with roads (edges), where distance represents FED, and then increasing certain speed limits. This fails because phones are not distributed uniformly; *Index Diachronica* has more attested occurrences of vowels than of consonants. So, the allegorical speed limits between "vowel cities" become so high that the paths between them act as "freeways." In analogous manner, the shortest paths between consonants tend to travel unnaturally through several vowels (e.g. k > g > w > u > o > a > ɛ > e > i > j > ʒ > d͡ʒ > t͡ʃ), in the same way that drivers may take a freeway to a neighboring city, even if the freeway entrance is not on the way. Our neural approach mitigates this by weighting the features of FED via probabilities between 0 and 1.

### 5.3 Sound laws from ASLI

Impressively, our method was able to infer high quality trees even when we replaced expert sound laws with ASLI, reproducing ~87.6% of expert *butterflies*. This indicates that the quality of sound laws generated by the minimal generalization learner may be sufficient for phylogenetic inference. Once again in this set of experiments, the 1-layer neural network performed best.

The sound laws produced by ASLI tend to have specific contexts. As mentioned in §4.2, we filter out generated sound laws with low accuracy, which tend to be laws with more general contexts. (These would be desirable if not for their lack of applicability to the data.) This leaves us with a large number of hyper-specific sound laws that only apply to a few daughter languages in the dataset. As illustrated in Figure 3, ASLI produces many more sound laws than are provided by the expert, but most of them apply only to a few languages. It seems that although these specific contexts may limit the potential shared innovations to be explored across languages, sufficiently useful information is still produced among the large number of sound laws produced. In other words, while an expert may provide optimal sound laws such that an accurate tree can be predicted with only a few dozen total, our ASLI approach produces enough—albeit lower quality—sound laws to yield similar performance. While the problem of insufficiently general sound laws may be due to our small dataset size, the lack of consideration for rule order could also play to ASLI's disadvantage, since minimal generalization was not originally designed to learn ordered changes.

## 6 Conclusion and Future Work

We propose a novel method to automatically predict intermediate sound changes for phylogenetic inference, via neural weighting of feature-based edit distance between phones. When we apply our method with a single-layer network, we accurately predict 88% of binary-branching language quartets (*butterflies*) in a gold Tukanoan phylogeny. Furthermore, our typologically informed neural approach based on articulatory features produces intermediate sound changes that capture expert intuitions on phonetic naturalness. Our analysis shows that not only does phonetic plausibility matter, but so does the accuracy of sound transition costs for successful phylogenetic inference. We also present a method to predict sound laws automatically via minimal generalization, which creates less generalizable sound laws than the expert but nonetheless provides sufficient phylogenetic signal to approach the performance of using expert sound laws. Combined, these methods can significantly reduce the cost, in terms of time and human expert annotation, of phylogenetic inference.

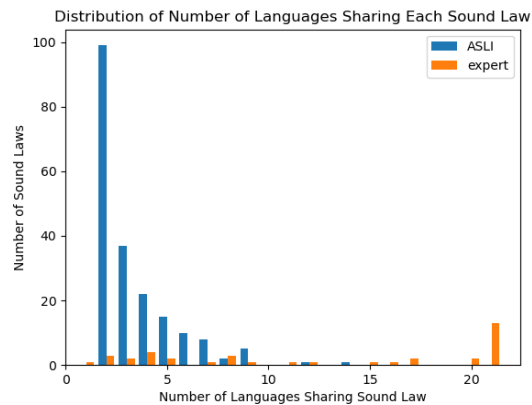

Figure 3: Histogram showing distribution of number of languages sharing sound laws, given by the expert and ASLI. ASLI produces many more sound laws, but they are shared between fewer languages.

Future work in this vein may involve exploration of other ASLI approaches (List, 2019; Luo, 2021), pruning the phone graph using PHOIBLE to focus on cross-linguistically frequent phonemes (Moran et al., 2014), generalization of our approach to other language families such as Polynesian, and incorporation of MCMC methods to jointly reconstruct protoforms and phylogenies.

## Limitations

Distinctive feature theory does not take some aspects of acoustic similarity into account. For instance, the common sound change p > ʔ is mo-tivated by acoustic factors, such as [p] having a weak burst. In addition, Schweikhard and List (2020) caution that *Index Diachronica* (Anonymous, 2016) does not always cite reliable sources. As such, we may wish to decrease the effect of our TYPOLOGICAL GROUNDING and instead include more language family-specific information a linguist would have just by analyzing its phoneme inventory. Historical linguists actually value rare sound changes that regularly occur since they are less likely to be parallel innovations and thus provide phylogenetic signal. The sound changes on which we train are only substitutions, so our DFWED may not handle insertions or deletions well. Additionally, none of the methods we mention handle borrowing or parallel innovations (homoplasy in Chacon and List (2016)), which means the methods here would not generalize well for Chinese and Romance. Furthermore, our baseline involves a different maximum parsimony method (Wagner parsimony) than Chacon and List (2016)'s modified Sankoff parsimony, which muddies the comparison between the two. That our shared innovations baseline outperforms the cognacy baseline cannot actually tell us that shared innovations outperforms cognacy information in general, because our gold tree was generated in part using shared innovations. Another limitation is that the genetic search algorithm does not scale well with more languages. Finally, the rules we learn in our minimal generalization learner also do not consider the relative chronology of the sound laws, as a historical phonologist would.

## Acknowledgement

We thank Maggie Huang for her independent study on phylogenetic inference that informed this work, Gašper Beguš for directing us to KonsonantalWandel (Kümmel, 2007), Edwin Ko for sharing his expertise in phylogenetic inference, Abhishek Vijayakumar for providing feedback on our initial phone graph ideas and for proofreading our paper, Melinda Fricke for answering our questions regarding the acoustic motivations behind sound change, and Leon Lu for proofreading our paper.

This material is based on research sponsored in part by the Air Force Research Laboratory under agreement number FA8750-19-2-0200. The U.S. Government is authorized to reproduce and distribute reprints for Governmental purposes notwithstanding any copyright notation thereon. The views and conclusions contained herein are those of the authors and should not be interpreted as necessarily representing the official policies or endorsements, either expressed or implied, of the Air Force Research Laboratory or the U.S. Government.

## A  Experimental Details

### A.1  Neural network hyperparameters

- num_epochs = 25
- batch_size = 5
- optimizer = Adam
- learning_rate = 0.001
- train_test_split = 0.9
- seed = 411

## B  Best tree analysis

Our algorithm correctly groups Western Tukano (`Kue`, `Kor`, `Mai`, `Sek`, and `Sio`) varieties in the same branch, with `Kue`/`Kor` and `Sio`/`Sek` paired (and `Mai` by itself), correctly (though it does not predict the correct chronology of the branching). "East-Eastern Tukano" varieties (`Tuk`, `Wan`, `Pir`, `Tuy`, `Yur`, `Pis`, `Kar`, `Tat`, and `Bar`) are also grouped in the same branch, with `Tuk` correctly splitting off the earliest and `Pir`/`Wan` and `Yur`/`Tuy` paired correctly. However, `Pis`, `Kar`, `Bar`, and `Tat` are predicted in an incorrect order. As for "West-Eastern" Tukano (`Bas`, `Mak`, `Yup`, `Des`, and `Sir`), our tree's grouping is wrong: `Des`/`Sir` and `Yup` are correct relative to each other but are in the wrong branch, `Mak`/`Bas`, `Tan`, and `Kub` are grouped correctly, but our predicted tree splits `Kub` and `Tan`, while the gold tree does not. Refer to Chacon and List (2016) for the original names of each variety.

## C  Example sound laws

Examples of sound laws from our ASLI method:

- sufficiently general: e →ẽ / (n|m) __
- too specific: p →m / (#) (pˀ|ˀp) (o) __ (a) (#)

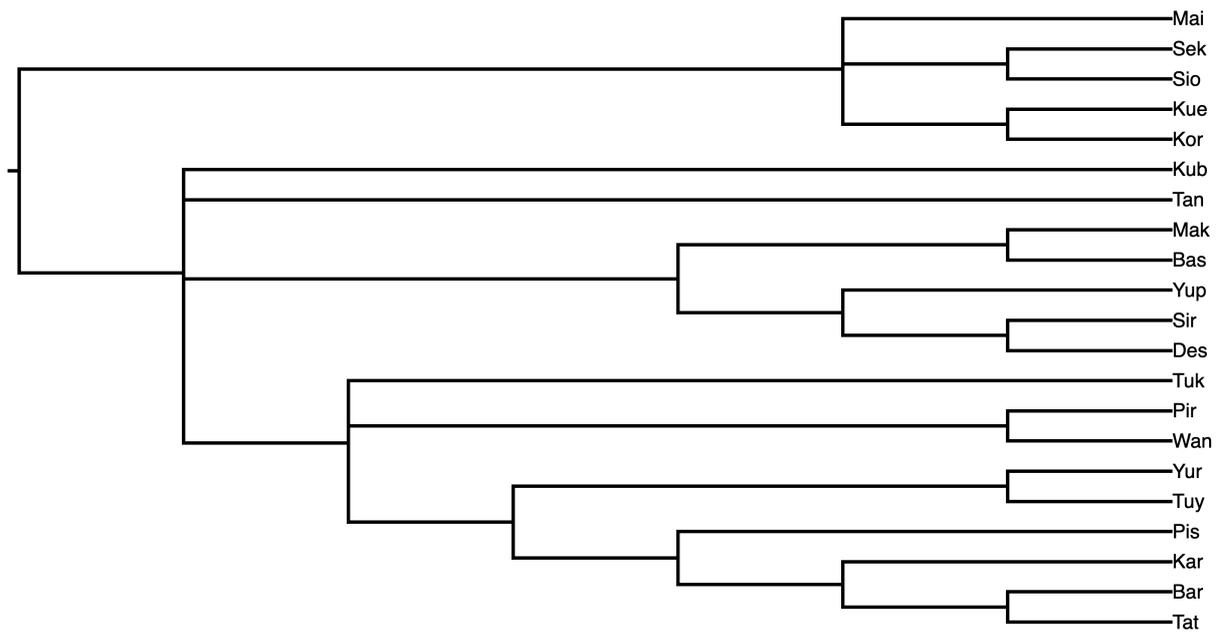

Figure 4: Gold Tukanoan phylogeny from Chacon and List (2016), which is a consensus of Chacon (2014) and their DiWeST tree

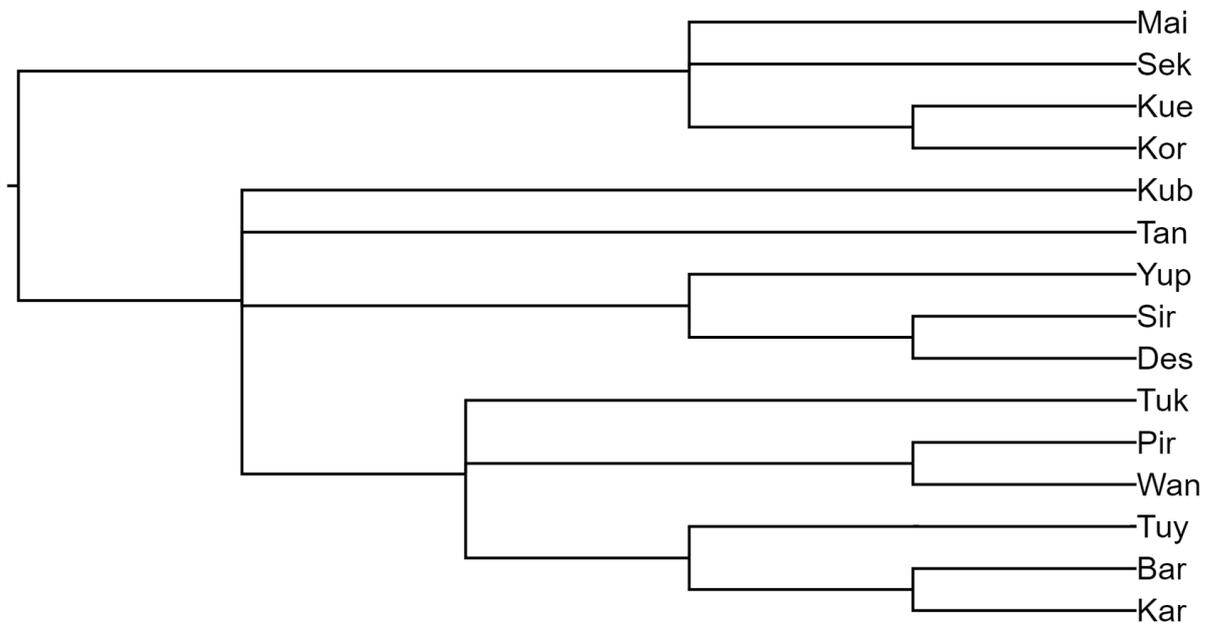

Figure 5: Gold Tukanoan phylogeny from Chacon and List (2016) but with only the 15 varieties in Chacon (2014)

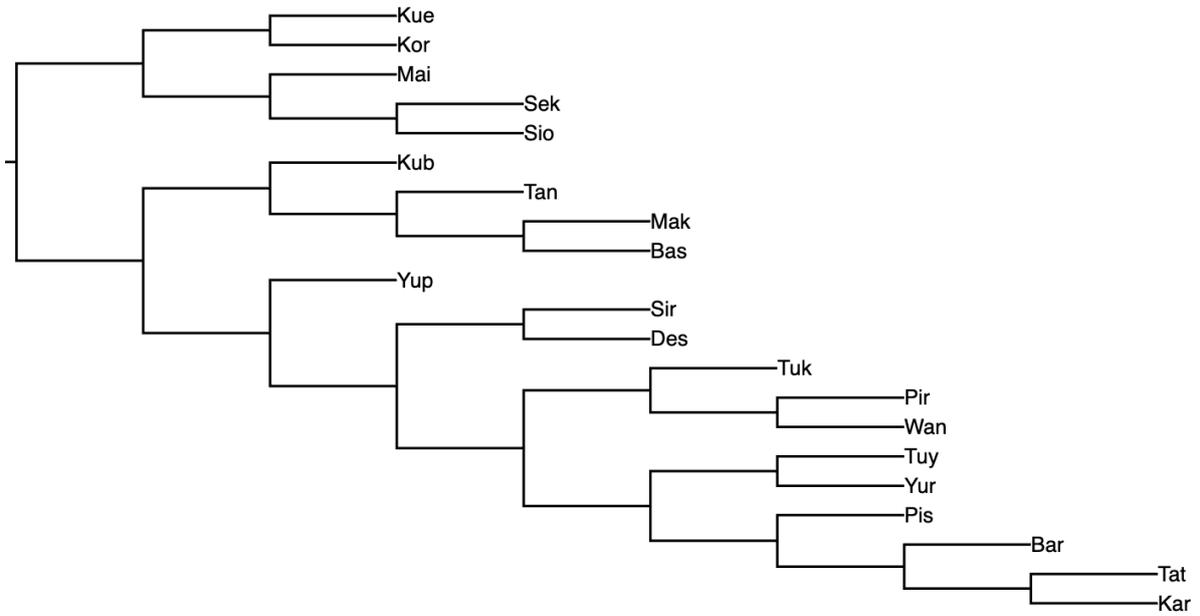

Figure 6: The predicted tree with the lowest GQD when compared to the gold tree (Figure 4), generated from the main experiments with a 1 layer neural network and using expert sound laws

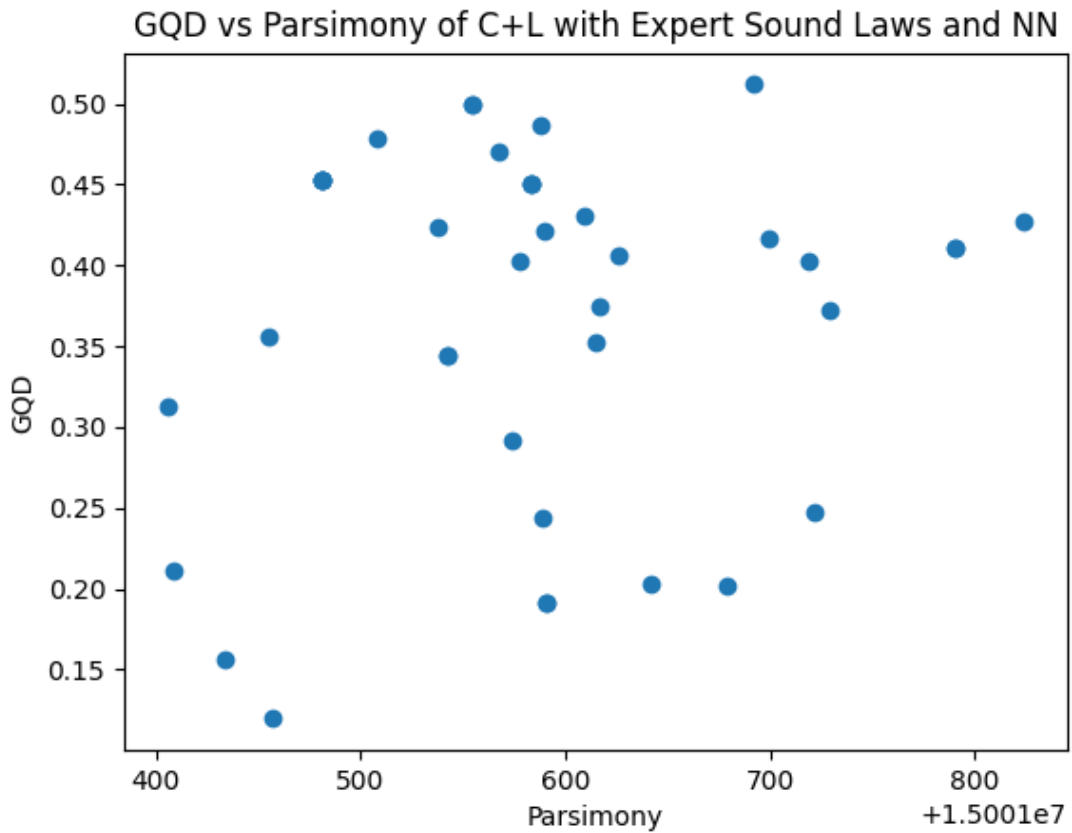

Figure 7: GQD vs parsimony of all 40 runs of C+L with expert sound laws and various NNs. The Spearman's coefficient between GQD and parsimony is -0.0373 ($p = 0.819$).

| | Experiment | Shortest paths (corr. 6) | Shortest paths (corr. 13) | Avg. Num. Paths | Avg. Num. Edges/Path | Recall |
|---|---|---|---|---|---|---|
| 3 | C+L, w/ AISCP (*FED* ablation) | k > k͡x > t͡ɕ > t͡ʃ | p > h | 1.78 | 3.47 | 0.522 |
| 5 | C+L, w/ AISCP, 1 layer NN | k > c > t͡ɕ > t͡ʃ | p > f > h | 1.0 | 2.10 | 0.342 |
| 6 | C+L, w/ AISCP, 4 layer NN | k > c > t͡ʃ | p > h | 1.0 | 1.94 | 0.366 |
| 7 | C+L, w/ AISCP, 8 layer NN | k > c > tʲ > t͡ʃ | p > f > h | 1.0 | 2.16 | 0.354 |
| 8 | C+L, w/ AISCP, 16 layer NN | k > c > t͡ʃ | p > f > h | 1.0 | 2.01 | 0.329 |
| | gold, expert sound laws | k > kʲ > t͡ʃ, k > kʰ > t͡ʃ | p > *pʰ > *ɸ > h | 1.31 | 1.86 | - |

Table 2: Comparison of the intermediate sound changes predicted in our main experiments using our DWFED method, with the unweighted ablation and the expert's posited sound changes included for comparison. Corr. 6 and 13 each refer to the index of the sound correspondence in Chacon and List (2016)'s dataset of annotated sound correspondences. Avg. # Paths refers to the average number of unique intermediate paths between proto-sound and reflex pairs in the dataset. Avg. # Edges/Path denotes the average number of edges in a shortest path. For Chacon and List's expert sound laws, we consider all paths in the calculation, since their paths are unweighted. Recall is the number of phones in the expert's proposed intermediate sound changes that appear in our predicted sound correspondences.